\documentclass{sigchi}

\doi{}
\isbn{}

\usepackage{balance}       
\usepackage{graphics}      
\usepackage[T1]{fontenc}   
\usepackage{txfonts}
\usepackage[pdflang={en-US},pdftex]{hyperref}
\usepackage{color}
\usepackage{booktabs}
\usepackage{textcomp}

\usepackage{microtype}        
\usepackage{ccicons}          

\usepackage{todonotes}

\def\plaintitle{Addressing Two Problems in Deep Knowledge Tracing via Prediction-Consistent Regularization}

\def\emptyauthor{}
\def\plainkeywords{Knowledge tracing; Deep learning; Regularization; Educational data mining; Personalized learning; Sequence modeling}

\makeatletter
\def\url@leostyle{%
  \@ifundefined{selectfont}{
    \def\UrlFont{\sf}
  }{
    \def\UrlFont{\small\bf\ttfamily}
  }}
\makeatother
\def\pprw{8.5in}
\def\pprh{11in}

\definecolor{linkColor}{RGB}{6,125,233}
\hypersetup{%
  pdftitle={\plaintitle},
  pdfauthor={\emptyauthor},
  pdfkeywords={\plainkeywords},
  pdfdisplaydoctitle=true, 
  bookmarksnumbered,
  pdfstartview={FitH},
  colorlinks,
  citecolor=black,
  filecolor=black,
  linkcolor=black,
  urlcolor=linkColor,
  breaklinks=true,
  hypertexnames=false
}

\usepackage{hyperref}
\usepackage{url}

\usepackage{amsmath, amssymb, bm}
\usepackage{graphicx}
\usepackage{subcaption}
\usepackage{dblfloatfix}    
\graphicspath{ {images/} }

\usepackage{multirow}
\usepackage{adjustbox}
\usepackage{tabularx}

\usepackage[toc,page]{appendix}

\usepackage{amssymb}
\usepackage{bm}

\def \b {\mathbf{b}}
\def \c {\mathbf{c}}

\def \f {\mathbf{f}}

\def \h {\mathbf{h}}
\def \i {\mathbf{i}}

\def \o {\mathbf{o}}

\def \x {\mathbf{x}}
\def \y {\mathbf{y}}

\def \P {\mathbf{P}}

\def \S {\mathbf{S}}

\def \W {\mathbf{W}}
\def \X {\mathbf{X}}

\def \Lcal {\mathcal{L}}

\def \deltabm {\bm{\delta}}

\def \lr {\lambda_r}

\usepackage{soul}
\usepackage{color}

\begin{document}

\title{\plaintitle}

\numberofauthors{2}
\author{%
  \alignauthor{Chun-Kit Yeung\\
    \affaddr{Hong Kong University of Science and Technology}\\
    \email{ckyeungac@cse.ust.hk}}\\
  \alignauthor{Dit-Yan Yeung\\
    \affaddr{Hong Kong University of Science and Technology}\\
    \email{dyyeung@cse.ust.hk}}\\
}

\maketitle

\begin{abstract}
Knowledge tracing is one of the key research areas for empowering personalized education. It is a task to model students' mastery level of a knowledge component (KC) based on their historical learning trajectories. In recent years, a recurrent neural network model called deep knowledge tracing (DKT) has been proposed to handle the knowledge tracing task and literature has shown that DKT generally outperforms traditional methods. However, through our extensive experimentation, we have noticed two major problems in the DKT model. The first problem is that the model fails to reconstruct the observed input. As a result, even when a student performs well on a KC, the prediction of that KC's mastery level decreases instead, and vice versa. Second, the predicted performance for KCs across time-steps is not consistent. This is undesirable and unreasonable because student's performance is expected to transit gradually over time. To address these problems, we introduce regularization terms that correspond to \emph{reconstruction} and \textit{waviness} to the loss function of the original DKT model to enhance the consistency in prediction. Experiments show that the regularized loss function effectively alleviates the two problems without degrading the original task of DKT.\footnote{The implementation of this work is available on \url{https://github.com/ckyeungac/deep-knowledge-tracing-plus}.}
\end{abstract}



\section{Introduction}

\begin{figure*}[ht]
	\begin{center}
		\includegraphics[width=0.90\linewidth]{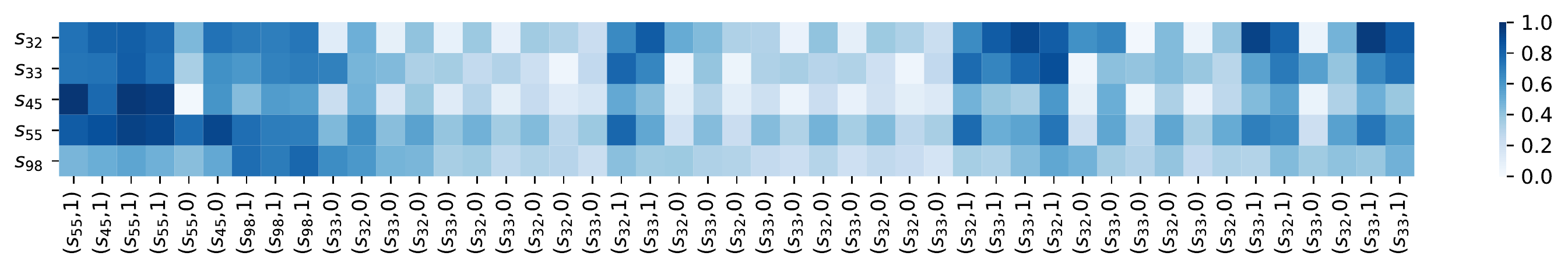}
	\end{center}
	\caption{A student, namely id-1, from the ASSISTment 2009 dataset is used to plot the heatmap illustrating the two problems in DKT. The label $s_i$ in the vertical dimension corresponds to the skill tag, and only those which have been answered by the student are shown. The label in the horizontal dimension refers to the input fed into the DKT at each time-step. The color of the heatmap indicates the predicted probability that the student will answer $s_i$ correctly in the next time-step, i.e., $p(a_{t+1} = 1~|~q_{t+1} = s_i)$. The darker the color, the higher the probability.}
	\label{fig:weaknessOfDKT}
\end{figure*}


With the advancement of digital technologies, online platforms for intelligent tutoring systems (ITSs) and massive open online courses (MOOCs) are becoming prevalent. These platforms produce massive datasets of student learning trajectories about the \emph{knowledge components} (KCs), where KC is a generic term for skill, concept, exercise, etc. The availability of online activity logs of students has accelerated the development of learning analytics and educational data mining tools for predicting the performance and advising the learning of students. Among many topics, knowledge tracing (KT) is considered to be important for enhancing personalized learning. KT is the task of modeling student's \emph{knowledge state}, which represents the mastery level of KCs, based on historical data.
With the estimated students' knowledge state, teachers or tutors can gain a better understanding of the attainment levels of their students and can tailor the learning materials accordingly. Moreover, students may also take advantage of the learning analytics tools to come up with better learning plans to deal with their weaknesses and maximize their learning efficacy.

Generally, the KT task can be formalized as follows: given a student's historical interactions $\X_t = ( \x_{1}, \x_{2}, ..., \x_{t} )$ up to time $t$ on a particular learning task, it predicts some aspects of his next interaction $\x_{t+1}$. Question-and-answer interactions are the most common type in KT, and thus $\x_t$ is usually represented as an ordered pair $(q_t, a_t)$ which constitutes a tag for the question $q_t$ being answered at time $t$ and an answer label $a_t$ indicating whether the question has been answered correctly. In many cases, KT usually seeks to predict the probability that the student will answer the question correctly during the next time-step, i.e., $p(a_{t+1} = 1 | q_{t+1}, \X_t)$. Many approaches have been developed to solve the KT problem, such as using the hidden Markov model~(HMM) in Bayesian knowledge tracing (BKT)~\cite{UMUAI1994_Corbett_BKT} and the logistic regression model in performance factors analysis (PFA)~\cite{AIE2009_Pavlik_PFA}. More recently, a recurrent neural network (RNN) model has been applied in a method called deep knowledge tracing (DKT)~\cite{NIPS2015_Piech_DKT}. Experiments show that DKT outperforms traditional methods without requiring substantial feature engineering by humans.

Although DKT achieves impressive performance for the KT task, we have noticed two major problems in the prediction results of DKT when trying to replicate the experiments in~\cite{NIPS2015_Piech_DKT} (where the authors adopted the skill-level tag as the question tag.) These two problems are illustrated using a heatmap in Figure~\ref{fig:weaknessOfDKT}, which visualizes the predicted knowledge state at each time-step of a student (namely id-1) from the ASSISTment 2009 skill builder dataset. 

The first problem of DKT is that the DKT model fails to reconstruct the input information in prediction. When a student performs well in a learning task related to a skill $s_i$, the model's predicted performance for the skill $s_i$ may drop instead, and vice versa. For example, at the $6^{th}$ time-step in Figure~\ref{fig:weaknessOfDKT}, the probability of correctly answering the exercise related to $s_{45}$ increases compared to the previous time-step even though the student answered $s_{45}$ incorrectly.

Second, it is observed that the transition in prediction outputs, i.e., the student's knowledge states, across time-steps is not consistent. As depicted in Figure~\ref{fig:weaknessOfDKT}, there are sudden surges and plunges in the predicted performance of some skills across time-steps. For example, the probabilities of correctly answering $s_{32}$,  $s_{33}$, $s_{45}$, and $s_{55}$ fluctuate when the student answers $s_{32}$ and $s_{33}$ in the middle of the learning sequence. This is intuitively undesirable and unreasonable as students' knowledge state is expected to transit gradually over time, but not to alternate between mastered and not-yet-mastered. Such wavy transitions are therefore not favorable as it would mislead the interpretation of the student's knowledge state.

To address the problems described above, we propose to augment the original loss function of the DKT model by introducing additional quality measures other than the original one which solely considers the prediction accuracy of the next interaction. Specifically, we define the \emph{reconstruction} error and \emph{waviness} measures and use them to augment the original loss function as a regularized loss function. Experiments show that the regularized DKT is more accurate in reconstructing the answer label of the observed input and is more consistent in its prediction across time-steps, yet without sacrificing the prediction accuracy for the next interaction.

Our main contributions are summarized as follows:
\begin{itemize}
	\item Two problems in DKT that have not been revealed in the literature are raised: failure in current observation reconstruction and wavy prediction transition;
	\item Three regularization terms for enhancing the consistency of prediction in DKT are proposed: $r$ to address the reconstruction problem, and $w_1$ and $w_2$ to address the wavy prediction transition problem;
	\item Five other performance measures are proposed to evaluate three aspects of goodness in KT: AUC(C) for the prediction performance of the current interaction, $w_1$ and $w_2$ for the waviness in KT's prediction overall, and $m_1$ and $m_2$ for the consistency between the current observation and the corresponding change in prediction.
\end{itemize}

\section{Background} \label{backgound}
\begin{figure*}[!b]
	\centering
    \includegraphics[width=0.8\linewidth]{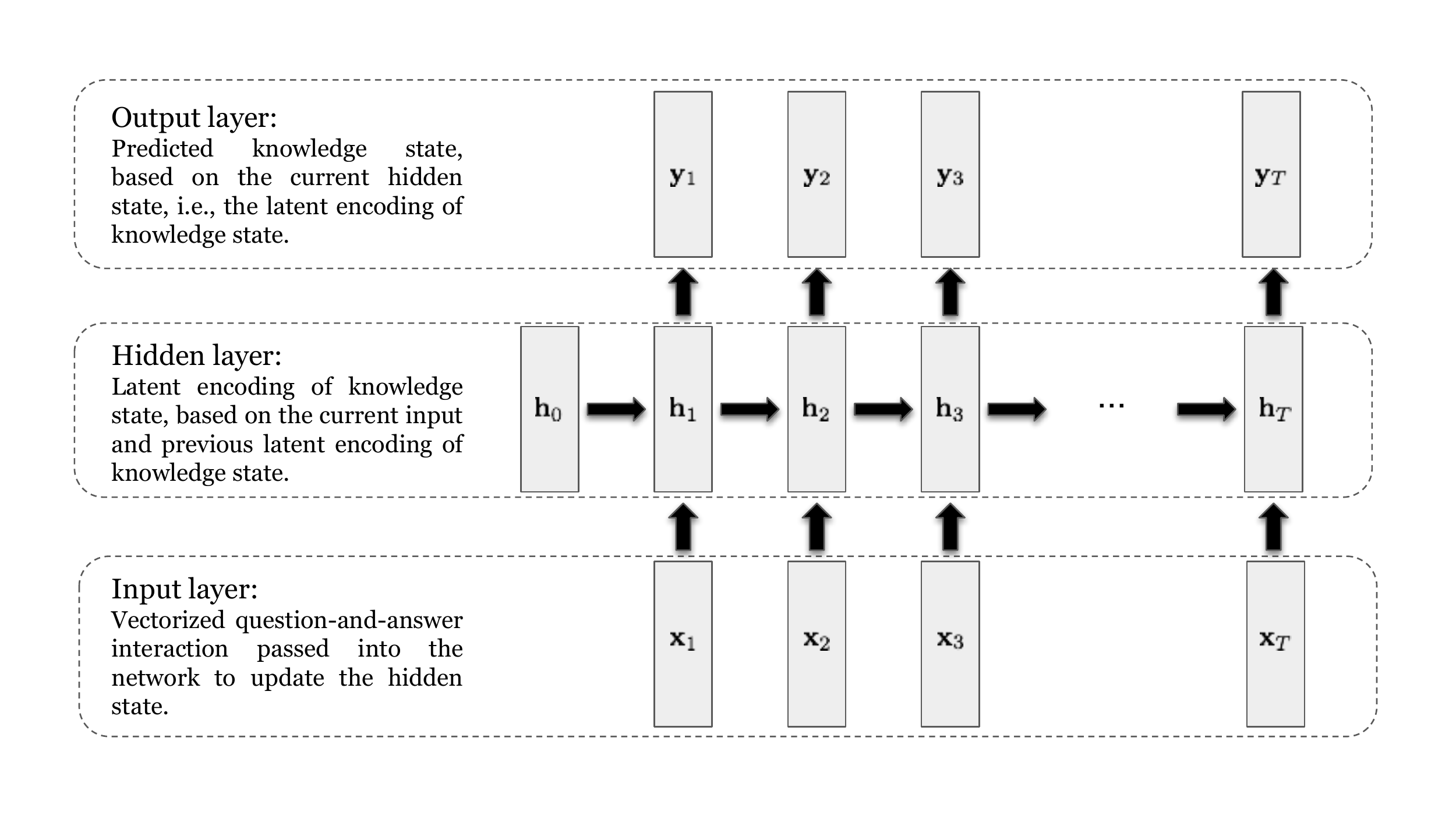}
    \caption{Unfolded version of the RNN architecture for DKT. The hidden state is processed differently in the vanilla RNN (eq.~\eqref{rnn_hidden_state}) or LSTM-RNN (eq.~\eqref{lstm_hidden_state}). $\h_0$ is the initial hidden state in the RNN, and it is usually initialized randomly or to a zero vector.}
    \label{fig:rnn_architecture}
\end{figure*}

Researchers have been investigating mathematical and computational models to tackle the KT task since the 1990s. Various approaches, ranging from probabilistic models to deep neural networks, have been developed over the past two decades.

The Bayesian knowledge tracing (BKT) model was proposed in~\cite{UMUAI1994_Corbett_BKT} during the 1990s. It models the knowledge states of KCs using one HMM for each KC. Specifically, the hidden state in the HMM represents the student's knowledge state which indicates whether or not the KC is mastered. However, many simplifying assumptions adopted by BKT are unrealistic. For example, BKT assumes that forgetting does not occur and the KCs are mutually independent. To address these shortcomings, some variants of BKT such as those with forgetting power~\cite{ITS2014_hawkins_learning} and skill dependency~\cite{IEEE2017_Kaesor_DynamicBKT} have been proposed. Extensive works have also been conducted to empower the individualization of BKT on both skill-specific parameters~\cite{UMAP2010_Pardos_BKT, UMAP2011_Pardos_KT} and student-specific parameters~\cite{IAIED2013_Yudelson_individualized}. Some other attempts have been made to extend the capabilities of BKT on the partial score~\cite{IAIED2013_Wang_Extending}, sub-skills and temporal features~\cite{EDM2014_Gonzales_General}, and more features from cognitive science such as the recency effect and contextualized trial sequences~\cite{EDM2016_Khajah_How}. However, it should be noted that such extensions often require considerable feature engineering efforts and may incur a significant increase in the computational requirements.

In the 2000s, learning factors analysis (LFA)~\cite{ITS2006_Cen_LFA} was proposed to model student knowledge states using the logistic regression model to deal with the multi-KCs issue and to incorporate student ability into the model. There is a reconfiguration of the LFA, called performance factors analysis (PFA)~\cite{AIE2009_Pavlik_PFA}, which offers higher sensitivity to student performance rather than student ability. Both LFA and PFA exploit the number of successes or failures of applying a KC to predict whether a student has acquired the knowledge about the KC. Although both LFA and PFA can handle a learning task that is associated with multiple KCs, they cannot deal with the inherent dependency among KCs, e.g., ``addition'' is a prerequisite of ``multiplication''. Moreover, the features used in LFA and PFA are relatively simple and they cannot provide a deep insight into students' latent knowledge state.

Recently, with a surge of interest in deep learning models, DKT~\cite{NIPS2015_Piech_DKT}, which models student's knowledge state based on an RNN, has been shown to outperform the traditional models, such as BKT and PFA, without the need for human-engineered features such as recency effect, contextualized trial sequence, inter-skill relationship, and students' ability variation~\cite{EDM2016_Khajah_How}. Since the DKT was proposed, a few comprehensive studies have been reported to compare DKT with other KT models~\cite{EDM2016_Wilson_IRT,EDM2016_Xiong_Going} or to apply the ideas of DKT to other applications~\cite{L@S2016_Reddy_LSE,EDM2017_Wang_Programming,arxiv2016_Steven_Modeling}. Nevertheless, to the best of our knowledge, all such attempts in the literature evaluate the DKT model mainly with respect to the prediction of the next interaction based on the \emph{area under the ROC curve} (AUC) measure, without considering other quality aspects in the prediction result.

\subsection{Review of Deep Knowledge Tracing}

\subsubsection{Recurrent Neural Networks}
DKT employs the RNN as its backbone model (see Figure~\ref{fig:rnn_architecture}). A (vanilla) RNN~\cite{ARXIV2015_Lipton_RNN} aims to map an input sequence $(\x_{1}, \x_{2}, \ldots, \x_{T})$ to an output sequence $(\y_{1}, \y_{2}, \ldots, \y_{T})$. To map the input to the output, the input vector undergoes a series of transformations via a hidden layer, which captures useful information that is hard to human-engineer, and forms a sequence of hidden states $(\h_{{1}}, \h_{{2}}, \ldots, \h_{{T}})$. More concretely, at time-step $t$, the hidden state $\h_{{t}}$ is the encoding of the past information obtained up to time-step $t-1$, i.e., $\h_{{t-1}}$, and the current input $\x_{{t}}$. The input-to-hidden transformation and hidden-to-output transformation can be stated mathematically as follows:
\begin{align}
\h_{{t}} &= \tanh( \W_{{hx}} \x_{t} + \W_{{hh}} \h_{{t-1}} + \b_{{h}} ) \label{rnn_hidden_state} \\
\y_{{t}} &= \sigma ( \W_{{hy}} \h_{{t}} + \b_{{y}} )\label{rnn_output}
\end{align}
where both the hyperbolic tangent $\tanh(\cdot)$ and the sigmoid function $\sigma(\cdot)$ are applied in an element-wise manner. The model is parameterized by a weight matrix $\W$ and a bias vector $\b$ with appropriate dimensions. 

Piech et al.~\cite{NIPS2015_Piech_DKT} adopts an RNN variant with long short-term memory (LSTM) cells. An LSTM cell incorporates three gates to imitate the human memory system~\cite{olah2015understanding} so as to calculate the hidden state $\h_t$. The three gates are forget gate $\f_{t}$, input gate $\i_{t}$ and output gate $\o_{t}$, which control a memory cell state~$\c_t$. Mathematically, they are simply three vectors calculated based on the current input $\x_{{t}}$ and the previous hidden state $\h_{{t-1}}$:
\begin{align}
  \f_t &= \sigma ( \W_{{f}} [\x_{{t}}, \h_{{t-1}}] + \b_{{f}} ), \nonumber \\
  \i_t &= \sigma ( \W_{{i}} [\x_{{t}}, \h_{{t-1}}] + \b_{{i}} ), \nonumber \\
  \o_t &=  \sigma ( \W_{{o}} [\x_{{t}}, \h_{{t-1}}] + \b_{{o}} ), \nonumber
\end{align}
where $[\cdot]$ denotes concatenation. Different gates play different roles to control what information should be stored in $\c_{{t}}$. The forget gate $\f_{t}$ decides what information to forget from the previous memory cell state $\c_{{t-1}}$, while the input gate $\i_{t}$ decides what new information $\tilde{\c}_t$ should be added to the recent cell state $\c_{{t}}$. Thus, the recent cell state $\c_{{t}}$ depends on the previous cell state after forgetting and the new information added from the input gate. Eventually, the output gate $\o_{t}$ determines what information should be extracted from $\c_{{t}}$ to form the hidden state $\h_{t}$. These can be expressed mathematically as follows:
\begin{align}
  \tilde{\c}_t &= \tanh ( \W_{{c}} [\x_{{t}}, \h_{{t-1}}] + \b_{{c}} ), \nonumber  \\
  \c_t &= \f_{{t}} \otimes \c_{t-1} + \i_{{t}} \otimes \tilde{\c}_{t}, \nonumber \\
  \h_t &= \o_{{t}} \otimes \tanh (\c_{{t}}), \label{lstm_hidden_state}
\end{align}
where $\otimes$ denotes element-wise multiplication. This formulation empowers RNN to store information occurred in a distance history, and thus has a more powerful capability than the vanilla RNN. The unfolded RNN architecture is visualized in Figure~\ref{fig:rnn_architecture}, with a high-level interpretation of DKT.

\subsubsection{DKT Problem Formulation} \label{section:dkt formulation}
  To train a DKT model, an interaction $(q_t, a_t)$ needs to be transformed into a fixed-length input vector $\x_{{t}}$. As a question can be identified by a unique ID, it can be represented using one-hot encoding as a vector $\deltabm(q_{t})$. The corresponding answer label can also be represented as the one-hot vector $\deltabm(q_{t})$ of the corresponding question if a student answers it correctly, or a zero vector $\mathbf{0}$ otherwise. Therefore, if there are $M$ unique questions, then $\x_{{t}} \in \{ 0, 1\} ^{2M}$. 

After the transformation, DKT passes the $\x_t$ to the hidden layer and computes the hidden state $\h_{{t}}$ using the vanilla RNN or LSTM-RNN. As the hidden state summarizes the information from the past, the hidden state in the DKT can therefore be conceived as the latent knowledge state of student resulted from his past learning trajectory. This latent knowledge state is then disseminated to the output layer to compute the output vector $\y_{{t}}$, which represents the probabilities of answering each question correctly. For student $i$, if she has a sequence of question-and-answer interactions of length $T_{i}$, the DKT model maps the inputs $(\x_1^i, \x_2^i, \dots, \x_{T_i}^i)$ to the outputs $(\y_1^i, \y_2^i, \dots, \y_{T_i}^i)$ accordingly. 

The objective of the DKT is to predict the next interaction performance, so the target prediction will be extracted by performing a dot product of the output vector $\y_{t}^{i}$ and the one-hot encoded vector of the next question $\deltabm(q_{t+1}^{i})$. Based on the predicted output and the target output $a_{t+1}^{i}$, the loss function $\Lcal$ can be expressed as follows:
\begin{equation} \label{dkt_objective_function}
\mathcal{L} =  
\frac
{1}
{{\sum_{i=1}^{n} (T_i-1)}}
\left( \sum_{i=1}^{n} \sum_{t=1}^{T_i - 1} l\left( \y_{t}^{i} \cdot \deltabm(q_{t+1}^{i}), a_{t+1}^{i} \right) \right)
\end{equation}
where $n$ is the number of students and $l(\cdot)$ is the cross-entropy loss.

\section{Some Problems of DKT and their Remedies}

While we were replicating the experiments on the original DKT proposed in \cite{NIPS2015_Piech_DKT} using the skill builder dataset provided by ASSISTment in 2009 (denoted ASSIST2009)\footnote{More information about the dataset can be found in \url{https://sites.google.com/site/assistmentsdata/home/assistment-2009-2010-data}.}, we noticed two major problems in the prediction results of DKT. First, it sometimes fails to reconstruct the input observation because the prediction result is counterintuitive. When a student answers a question of skill $s_{i}$ correctly/incorrectly, the predicted probability of that student answering $s_{i}$ correctly sometimes decreases/increases instead. Second, the predicted knowledge state is wavy and inconsistent over time. This is not desirable because student's knowledge state is expected to transit only gradually and steadily over time. Therefore, we propose three regularization terms to rectify the consistency problem in the prediction of DKT: reconstruction error $r$ to resolve the reconstruction problem, and waviness measures $w_1$ and $w_2$ to smoothen the predicted knowledge state transition.

\subsection{Reconstruction Problem}
As we saw from Figure \ref{fig:weaknessOfDKT}, when the student answers $s_{32}$ incorrectly, the probability of correctly answering $s_{32}$ grows significantly compared to the previous time-step. This problem can be attributed to the loss function defined in the DKT model (eq.~\eqref{dkt_objective_function}). Specifically, the loss function takes only the predicted performance of the next interaction into account, but not the predicted performance of the current one. Accordingly, when the input order $\left( (s_{32}, 0), (s_{33}, 0) \right)$ occurs frequently enough, the DKT model will tend to learn that if a student answers $s_{32}$ incorrectly, he/she will likely answer $s_{33}$ incorrectly, but not $s_{32}$. Consequently, the prediction result is counterintuitive for the current observed input.

However, one might argue that such transition in prediction reveals that $s_{32}$ is a prerequisite of $s_{33}$.\footnote{We note that $s_{32}$ is ``Ordering Positive Decimals'' and $s_{33}$ is ``Ordering Fractions''.} This is because the predicted performance for $s_{32}$ is lower only when the DKT model receives $(s_{33}, 0)$, but it is higher when the DKT model receives $(s_{32}, 0)$. To gainsay the above argument, we are going to impeach by contradiction. We hypothesize that if $s_{32}$ is indeed a prerequisite of $s_{33}$, then when a student answers $s_{32}$ incorrectly in the current time-step, it is more probable that he/she will answer $s_{33}$ incorrectly in the next time-step, but not vice versa. To verify this hypothesis, Tables \ref{table:confusion_matrix_e32_e33} and \ref{table:confusion_matrix_e33_e32} are made to tabulate the frequency counts when $s_{32}$ and $s_{33}$ appear consecutively in different orders. According to the above hypothesis, it is expected that the lower right cell will have a larger value than the lower left cell in Table~\ref{table:confusion_matrix_e32_e33}, but not Table~\ref{table:confusion_matrix_e33_e32}.

\begin{table}[h]
		\begin{tabular}{l|l|c|c|c} 
			\multicolumn{2}{c}{}&\multicolumn{2}{c}{Next = $s_{33}$}&\\
			\cline{3-4}
			\multicolumn{2}{c|}{}&Correct&Incorrect&\multicolumn{1}{c}{Total}\\
			\cline{2-4}
			\multirow{2}{*}{Current = $s_{32}$}& Correct & 1543 & 159 & 1702 \\
			\cline{2-4}
			& Incorrect & 81 & 367 & 448\\
			\cline{2-4}
			\multicolumn{1}{c}{} & \multicolumn{1}{c}{Total} & \multicolumn{1}{c}{1624} & \multicolumn{1}{c}{526} & 				\multicolumn{1}{c}{2510}\\
		\end{tabular}
		\caption{Correctness matrix when the current skill is $s_{32}$ and the next skill is $s_{33}$.}
		\label{table:confusion_matrix_e32_e33}
        
        \vfill
        
		\begin{tabular}{l|l|c|c|c}
			\multicolumn{2}{c}{}&\multicolumn{2}{c}{Next = $s_{32}$}&\\
			\cline{3-4}
			\multicolumn{2}{c|}{}&Correct&Incorrect&\multicolumn{1}{c}{Total}\\
			\cline{2-4}
			\multirow{2}{*}{Current = $s_{33}$}& Correct & 1362 & 72 & 1434 \\
			\cline{2-4}
			& Incorrect & 90 & 361 & 451\\
			\cline{2-4}
			\multicolumn{1}{c}{} & \multicolumn{1}{c}{Total} & \multicolumn{1}{c}{1452} & \multicolumn{    1}{c}{433} & 				\multicolumn{1}{c}{1885}\\
		\end{tabular}
		\caption{Correctness matrix when the current skill is $s_{33}$ and the next skill is $s_{32}$.}
		\label{table:confusion_matrix_e33_e32}
\end{table}

From Table~\ref{table:confusion_matrix_e32_e33}, we can see that if a student answers $s_{32}$ incorrectly in the current time-step, it is more probable that he/she will answer $s_{33}$ incorrectly in the next time-step. However, Table~\ref{table:confusion_matrix_e33_e32} shows that if a student answers $s_{33}$ incorrectly, it is also more probable that he/she will answer $s_{32}$ incorrectly in the next time-step. This means that an inverse dependency also exists and contradicts the above hypothesis, and therefore the statement that $s_{32}$ is a prerequisite of $s_{33}$ becomes questionable. Moreover, the distributions of these two matrices would advocate that $s_{32}$ and $s_{33}$ are likely to be interdependent and acquired simultaneously.

If $s_{32}$ is not a prerequisite of $s_{33}$ and they should be acquired at the same time, then there should be room for improvement to deal with cases like $s_{32}$ and $s_{33}$ in DKT. As mentioned above, the loss function considers only the predicted performance of the next interaction but ignores the current one. An immediate remedy to alleviate the problem is to regularize the DKT model by taking the loss between the prediction and the current interaction into consideration. By doing so, the model will adjust the prediction with respect to the current input. Thus, a regularization term for the reconstruction problem is defined as follows:
\begin{equation} \label{reconstruction_regularizer}
r = 
\frac
{1}
{{\sum_{i=1}^{n} (T_i - 1)}}
\left( \sum_{i=1}^{n} \sum_{t=1}^{T_i - 1} l\left( \y_{t}^{i} \cdot \deltabm(q_{t}^{i}), a_{t}^{i} \right) \right).
\end{equation}

\subsection{Wavy Transition in Prediction}

The second problem is the wavy transition in the student's predicted knowledge state. This problem may be attributed to the hidden state representation in the RNN. The hidden state $\h_{{t}}$ is determined by the previous hidden state $\h_{{t-1}}$ and the current input $\x_{t}$. It summarizes the student's latent knowledge state of all the exercises in one single hidden layer. Although it is difficult to explicate how the elements in the hidden layer influence the predicted performance of the KCs, it is plausible to confine the hidden state representation to be more invariant via regularization over the output layer.

We define two waviness measures $w_1$ and $w_2$ as regularization terms to smoothen the transition in prediction:
\begin{align}
w_1 &= \label{waviness_l1}
\frac
{\sum_{i=1}^{n} \sum_{t=1}^{T_i - 1} \|  \y^{i}_{t+1} -\y^{i}_{t} \|_{1}}
{M\sum_{i=1}^{n} (T_i - 1)}, 
\\  
w^2_2 &= \label{waviness_l2}
\frac
{\sum_{i=1}^{n} \sum_{t=1}^{T_i - 1} \|  \y^{i}_{t+1} -\y^{i}_{t} \|_{2}^{2}}
{ M\sum_{i=1}^{n} (T_i - 1)}
. 
\end{align}
To quantify how disparate the two prediction vectors are, both $L1$-norm and $L2$-norm are used to measure the difference between the prediction results $\y^{i}_{t}$ and $\y^{i}_{t+1}$. This is similar to the elastic net regularization \cite{JRSS2005_Zou_ElasticNet}. The two measures are averaged over the total number of input time-steps and the number of KCs $M$. Thus, the magnitude of $w_1$ would be seen as the average value change of each component in the output vector between $\y_t$ and $\y_{t+1}$. The larger the values of $w_1$ and $w_2$, the more inconsistent the transitions in the model.

In summary, the original loss function is augmented by incorporating three regularization terms to give the following regularized loss function:
\begin{equation} \label{objective_function_with_waviness}
\mathcal{L}' 
=  \mathcal{L} + \lambda_r r + \lambda_{w_1} w_1 + \lambda_{w_2} w_2^2 
\end{equation}
where $\lambda_r$, $\lambda_{w_1}$ and $\lambda_{w_2}$ are regularization parameters. By training with this new loss function, the DKT model is expected to address the reconstruction and wavy transition problems.

\section{Experiments}

\subsection{Implementation} 
\paragraph{Experiment settings} In the following experiments, 20\% of the data is used as a test set and the other 80\% is used as a training set. Furthermore, 5-fold cross-validation is applied on the training set to select the hyperparameter setting. The test set is used to evaluate the model, and also to perform early stopping~\cite{NN1998_Prehelt_EarlyStopping}.
The weights of the RNN are 
initialized randomly from a Gaussian distribution with zero mean and small variance. For fair comparison, we follow the hyperparameter setting in~\cite{NIPS2015_Piech_DKT} even though it may not be optimal. A single-layer RNN-LSTM with a state size of 200 is used as the basis of the DKT model. The learning rate and the dropout rate are set to 0.01 and 0.5, respectively. In addition, we also consistently set the norm clipping threshold to 3.0. Moreover, our preliminary experiment using ASSIST2009 shows that using the exercise tag as the question tag induces data sparsity and results in poor performance\footnote{An AUC of 0.73 if 26,668 exercise IDs are used; an AUC of 0.82 if 124 unique skill IDs are used.}, so we adopt the skill tag to be the question tag in the following experiment.

\paragraph{Hyperparameter search} We perform hyperparameter
search for the regularization parameters $\lambda_r$, $\lambda_{w_1}$ and $\lambda_{w_2}$. At first, each parameter is examined separately to identify a range of values giving good results according to some evaluation measures
to be explained later. The initial search ranges for $\lambda_r$, $\lambda_{w_1}$ and $\lambda_{w_2}$ are \{0, 0.25, 0.5, 1.0\}, \{0, 0.001, 0.003, 0.01, 0.03, 0.1, 0.3, 1.0, 3.0, 10.0\}, and \{0, 0.001, 0.003, 0.01, 0.03, 0.1, 0.3, 1.0, 3.0, 10.0, 30.0, 100.0\}, respectively.
After narrowing down the range of each parameter,
a grid search over combinations of $\lambda_r$, $\lambda_{w_1}$ and $\lambda_{w_2}$ is conducted. The final search ranges for $\lambda_r$, $\lambda_{w_1}$ and $\lambda_{w_2}$ are \{0, 0.05, 0.10, 0.15, 0.20, 0.25\}, \{0, 0.01, 0.03, 0.1, 0.3, 1.0\}, and \{0, 0.3, 1.0, 3.0, 10.0, 30.0, 100.0\}, respectively.

\begin{table*}[tp]
	\centering
	\begin{adjustbox}{max width=1.0\textwidth,center}
		\begin{tabular}{| l | r r r | r r r r r r|}
			\hline
			Dataset & $\lambda_r$ & $\lambda_{w_1}$ & $\lambda_{w_2}$ & AUC(N) & AUC(C) & $w_1$ & $w_2$ & $m_1$ & $m_2$\\
			\hline
			\hline
			\multirow{ 2}{*}{ASSIST2009}
			& 0.0 & 0.0 & 0.0 & 0.8212$\pm$0.00023 & 0.9044$\pm$0.00151 & 0.0830$\pm$0.00381 & 0.1279$\pm$0.00535 & 0.3002$\pm$0.01832 & 0.0156$\pm$0.00205\\ 
			& 0.1 & 0.003 & 3.0 & \textbf{0.8227$\pm$0.00041} &  \textbf{0.9625$\pm$0.00365} &  \textbf{0.0229$\pm$0.00022} &  \textbf{0.0491$\pm$0.00033} &  \textbf{0.4486$\pm$0.00427} &  \textbf{0.0573$\pm$0.00132}\\ 
			\hline
			\multirow{ 2}{*}{ASSIST2015}
			& 0.0 & 0.0 & 0.0 & 0.7365$\pm$0.00045 & 0.8846$\pm$0.00185 & 0.0282$\pm$0.00116 & 0.0414$\pm$0.00162 & 0.6208$\pm$0.00799 & 0.0476$\pm$0.00044\\ 
			& 0.05 & 0.03 & 3.0 &  \textbf{0.7371$\pm$0.00017} &  \textbf{0.9233$\pm$0.00180} &  \textbf{0.0124$\pm$0.00017} &  \textbf{0.0210$\pm$0.00018} &  \textbf{0.8122$\pm$0.00915} &  \textbf{0.0591$\pm$0.00029}\\
			\hline
			\multirow{ 2}{*}{ASSISTChall}
			& 0.0 & 0.0 & 0.0 &  \textbf{0.7343$\pm$0.00021} & 0.7109$\pm$0.00579 & 0.0690$\pm$0.00130 & 0.1045$\pm$0.00181 & 0.1151$\pm$0.00920 & -0.0055$\pm$0.00199\\ 
			& 0.1 & 0.3 & 3.0 & 0.7285$\pm$0.00024 &  \textbf{0.8570$\pm$0.00175} &  \textbf{0.0147$\pm$0.00053} &  \textbf{0.0301$\pm$0.00091} &  \textbf{0.3052$\pm$0.00729} &  \textbf{0.0441$\pm$0.00039}\\ 
			\hline
			\multirow{ 2}{*}{Statics2011}
			& 0.0 & 0.0 & 0.0 & 0.8159$\pm$0.00037 & 0.7404$\pm$0.00556 & 0.1358$\pm$0.00970 & 0.1849$\pm$0.01308 & -0.2590$\pm$0.01124 & -0.0658$\pm$0.00502\\ 
			& 0.20 & 1.0 & 30.0 & \textbf{0.8349$\pm$0.00029} & \textbf{0.9038$\pm$0.00431} & \textbf{0.0074$\pm$0.00023} & \textbf{0.0130$\pm$0.00016} & \textbf{0.4761$\pm$0.03587} & \textbf{0.0315$\pm$0.00303}\\
			\hline
			\multirow{ 2}{*}{Simulated-5}
			& 0.0 & 0.0 & 0.0 & 0.8255$\pm$0.00034 & 0.8642$\pm$0.00265 & 0.0426$\pm$0.00136 & 0.0588$\pm$0.00199 & -0.1512$\pm$0.02501 & -0.0134$\pm$0.00569\\
			& 0.20 & 0.001 & 10.0 & \textbf{0.8264$\pm$0.00061} & \textbf{0.9987$\pm$0.00081} & \textbf{0.0196$\pm$0.00013} & \textbf{0.0426$\pm$0.00164} & \textbf{0.9064$\pm$0.01948} & \textbf{0.1659$\pm$0.00830}\\
			\hline
		\end{tabular}
	\end{adjustbox}
	\caption{The average test results of the evaluation measures, as well as their standard deviations from 3 trials are reported for both the DKT (with $\lambda_r = \lambda_{w_1} = \lambda_{w_2} = 0.0$) and DKT+ models. The hyperparameter setting reported for DKT+ is selected according to the following procedure using 5-fold cross-validation: (1)~select the DKT+ models with a lower value of $w_1$ than DKT, and (2)~among them, select the DKT+ with the highest value of $\text{AUC(N)}+\text{AUC(C)}+m_1+m_2$.}
	\label{table:dkt_best_result}
\end{table*}

\paragraph{Evaluation Measures}
The performance of the DKT is customarily evaluated by AUC, which provides a robust metric for binary prediction evaluation. An AUC score of 0.5 indicates that the model performance is merely as good as random guess. In this paper, we report not only the AUC for the next performance prediction (named AUC(N) in this paper for clarity) which is tantamount to the evaluation in \cite{NIPS2015_Piech_DKT}, but also five other quantities with respect to the reconstruction accuracy and consistency of the input observation as well as the waviness of the prediction result.

For the reconstruction accuracy, the AUC for the current performance prediction (called AUC(C)) is used. With regard to the consistency in prediction of the input observation, two more quantities $m_1$ and $m_2$ are defined to measure the consistency between the observed input and the change in the corresponding prediction. For a single student $i$ at time $t$, we define 
\begin{align}
m_{1,t}^{i} &= (-1)^{1-a_t^i} \text{ sign}\left( (\y_t^i - \y_{t-1}^i) \cdot \deltabm(q_t^i) \right),
\\	
m_{2,t}^{i} &= (-1)^{1-a_t^i} \left((\y_t^i - \y_{t-1}^i) \cdot \deltabm(q_t^i) \right),
\end{align}
and
\begin{align}
m_1 &= \frac { \sum_{i=1}^{n} \sum_{t=2}^{T_i} m_{1,t}^{i} } {\sum_{i=1}^{n} (T_i - 1) },\\
m_2 &= \frac { \sum_{i=1}^{n} \sum_{t=2}^{T_i} m_{2,t}^{i} } {\sum_{i=1}^{n} (T_i - 1) }.
\end{align}
Accordingly, when the model gives a correct change in prediction with respect to the input, we will obtain positive values for $m_{1,t}^{i}$ and $m_{2,t}^{i}$. Otherwise negative values will be obtained. A positive value of $m_1$ indicates that more than half of the predictions comply with the input observations; a zero value implies that the model makes half of the predictions change in the right direction while another half change in a wrong direction; a negative value means that the model makes more than half of the predictions change in a wrong direction. A similar interpretation also holds for $m_2$ though it takes changes in both the direction and magnitude into account. Accordingly, the higher the values of $m_1$ and $m_2$ are, the better the model is from the perspective of consistency in prediction for the current observation.

Besides, the waviness 
measures $w_1$ and $w_2$ are also used as performance measures to quantify the consistency in prediction of the other KCs in the model. We deem that a good DKT model should achieve a high AUC score while maintaining a low waviness value. 

\subsection{Datasets}

\paragraph{ASSISTment 2009 (ASSIST2009)} This dataset is provided by the ASSISTments online tutoring platform and has been used in several papers for the evaluation of DKT models. Owing to the existence of duplicated records in the original dataset~\cite{EDM2016_Xiong_Going}, we have removed them before conducting our experiments. The resulting dataset contains 4,417 students with 328,291 question-answering interactions from 124 skills.
Some of the students in this dataset are used for visualizing the prediction result.

\paragraph{ASSISTment 2015 (ASSIST2015)} This dataset contains 19,917 student responses for 100 skills with a total of 708,631 question-answering interactions. Although it contains more interactions than ASSIST2009, the average number of records per skill and student is actually smaller due to a larger number of students. 

\paragraph{ASSISTment Challenge (ASSISTChall)} This dataset has been made available for the 2017 ASSISTments data mining competition. It is richer in terms of the average number of records per student as there are 686 students with 942,816 interactions and 102 skills.

\paragraph{Statics2011} This dataset is obtained from an engineering statics course with 189,927 interactions from 333 students and 1,223 exercise tags.
We have adopted the processed data provided by \cite{WWW2017_Zhang_DKVMN} with a train/test split of ratio 70:30, and exercise tags are used.

\paragraph{Simulated-5} Piech et al.~\cite{NIPS2015_Piech_DKT}
also simulated 2000 virtual students' answering trajectories
in which the exercises are drawn from five virtual concepts. Each student answers the same sequence of 50 exercises each of which has a single concept $k \in \{1,\ldots,5\}$ and a difficulty level $\beta$, with an assigned ability $\alpha_k$ of solving the task related to the skill $k$. The probability of a student answering an exercise correctly is defined based on the conventional item response theory as $p(\text{correct} | \alpha, \beta) = c + \frac{1 - c}{1 + \exp(\beta - \alpha)}$,
where $c$ denotes the probability of guessing it correctly and it is set to $0.25$.

\subsection{Results}
\begin{figure*}[!t]
	\includegraphics[width=\textwidth]{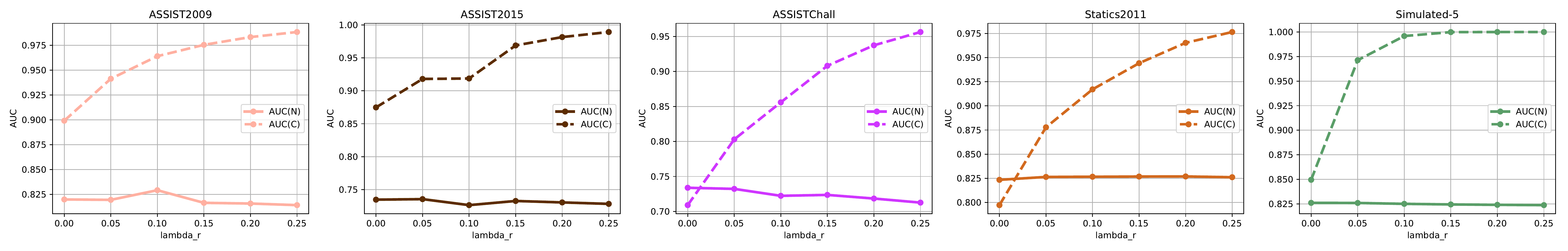}
	\caption{Average test AUC(N) and AUC(C) for different values of $\lambda_r$ over different combinations of $\lambda_{w_1}$ and $\lambda_{w_2}$.}
	\label{fig:dkt_lambda_r_on_aucc}
\end{figure*}

\begin{figure*}[!b]
	\centering
	\begin{subfigure}[b]{0.28\textwidth}
		\includegraphics[width=\textwidth]{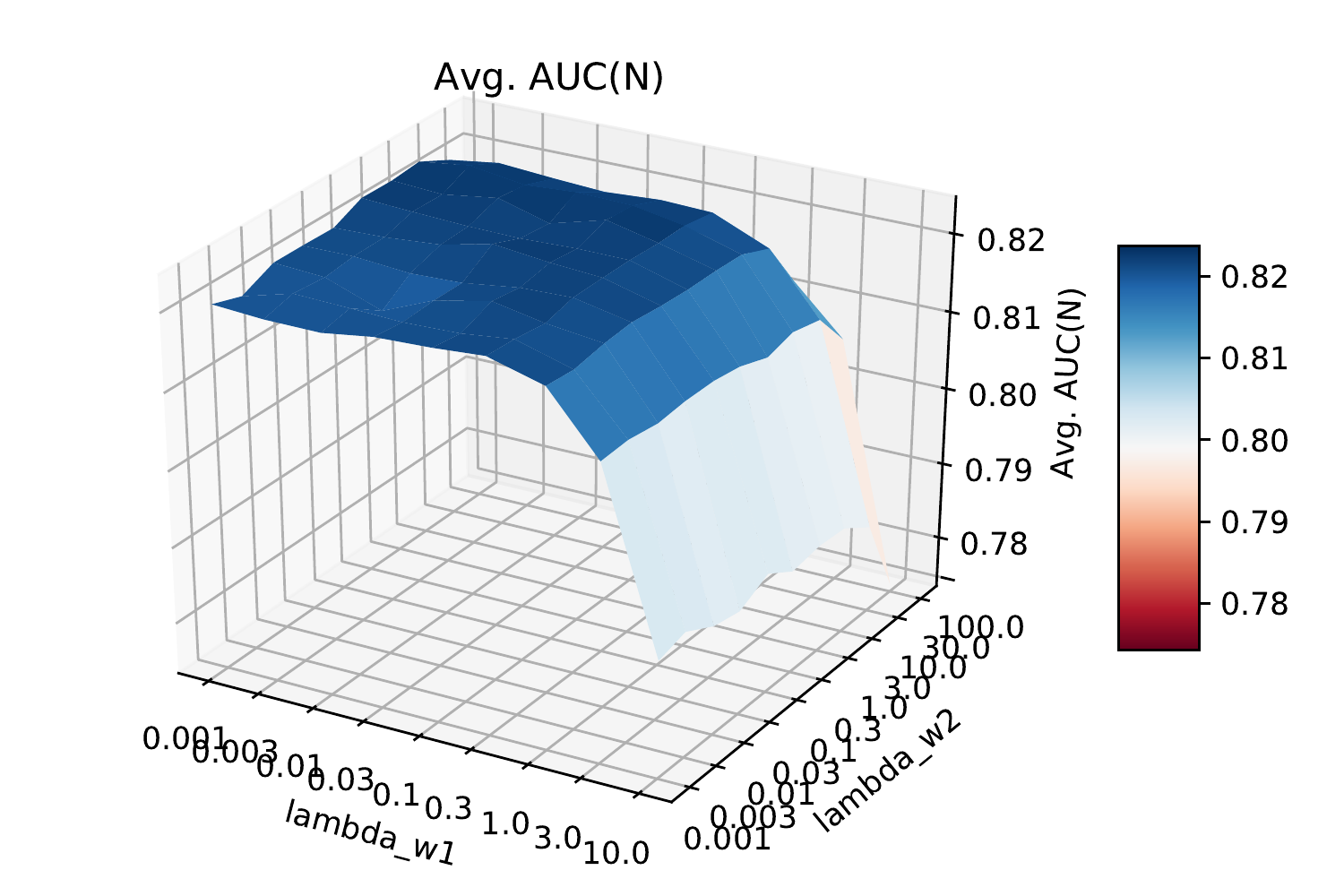}
	\end{subfigure}	
	~
	\begin{subfigure}[b]{0.28\textwidth}
		\includegraphics[width=\textwidth]{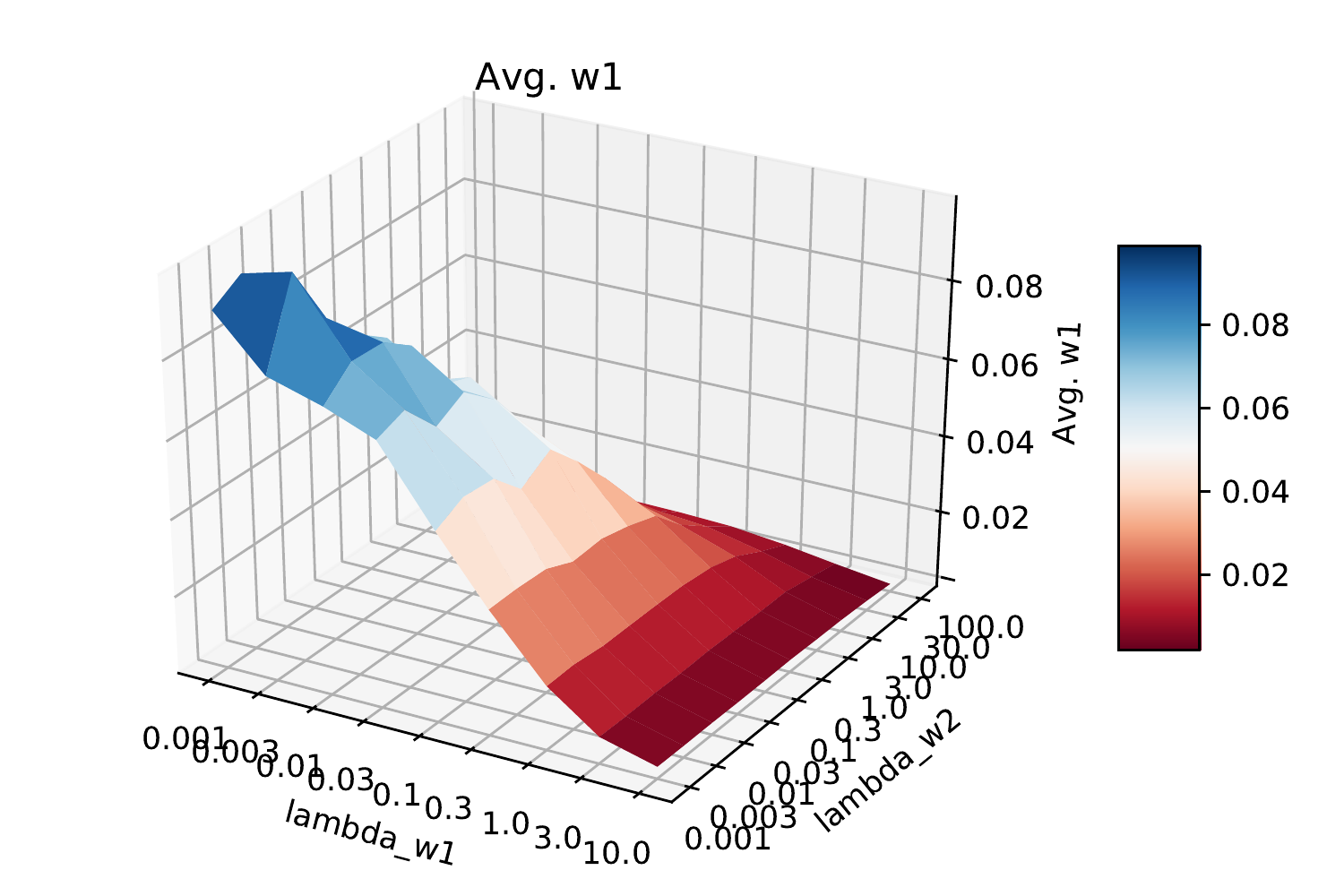}
	\end{subfigure}	
	~
	\begin{subfigure}[b]{0.28\textwidth}
		\includegraphics[width=\textwidth]{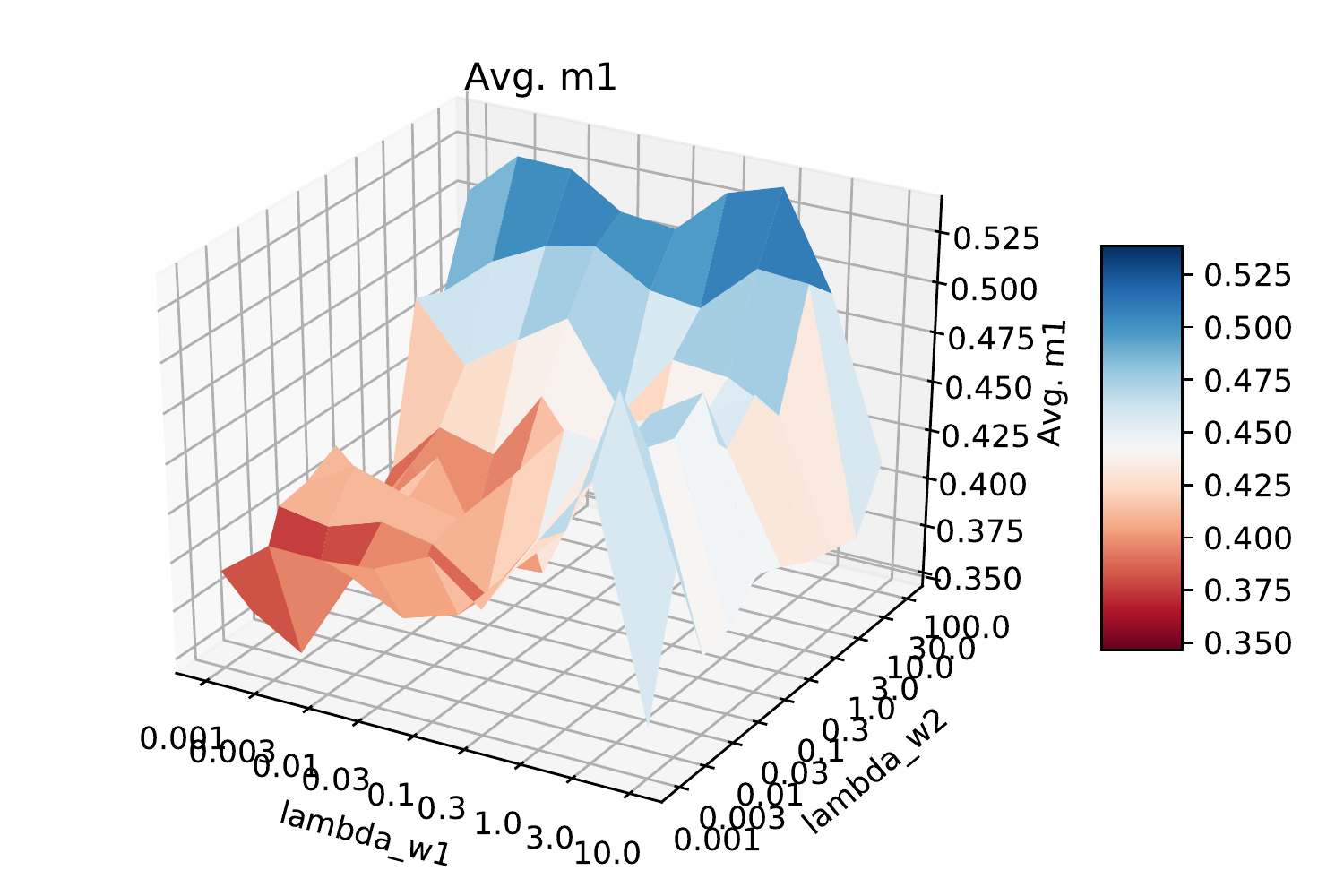}
	\end{subfigure}
	~
	\begin{subfigure}[b]{0.28\textwidth}
		\includegraphics[width=\textwidth]{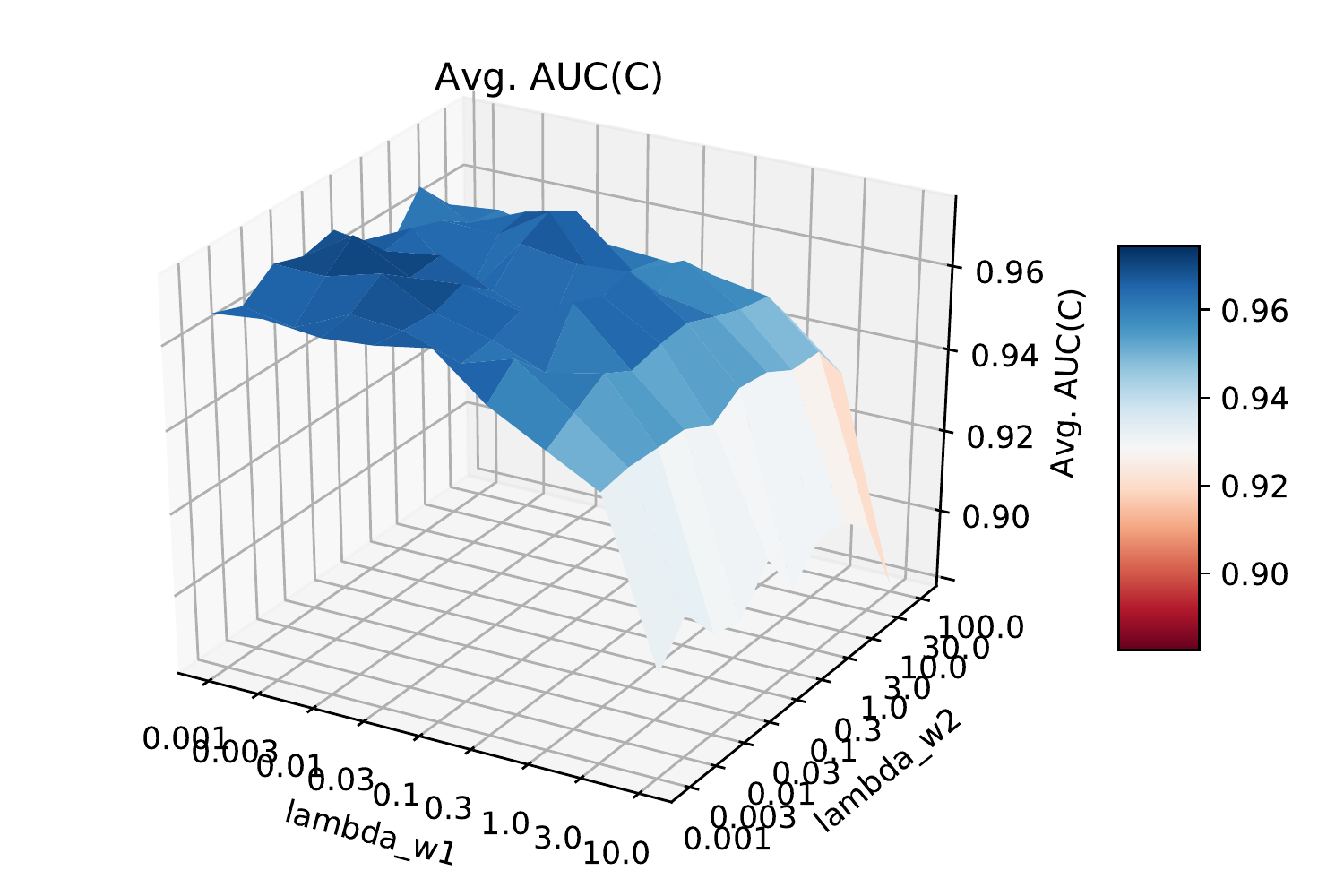}
	\end{subfigure}	
	~
	\begin{subfigure}[b]{0.28\textwidth}
		\includegraphics[width=\textwidth]{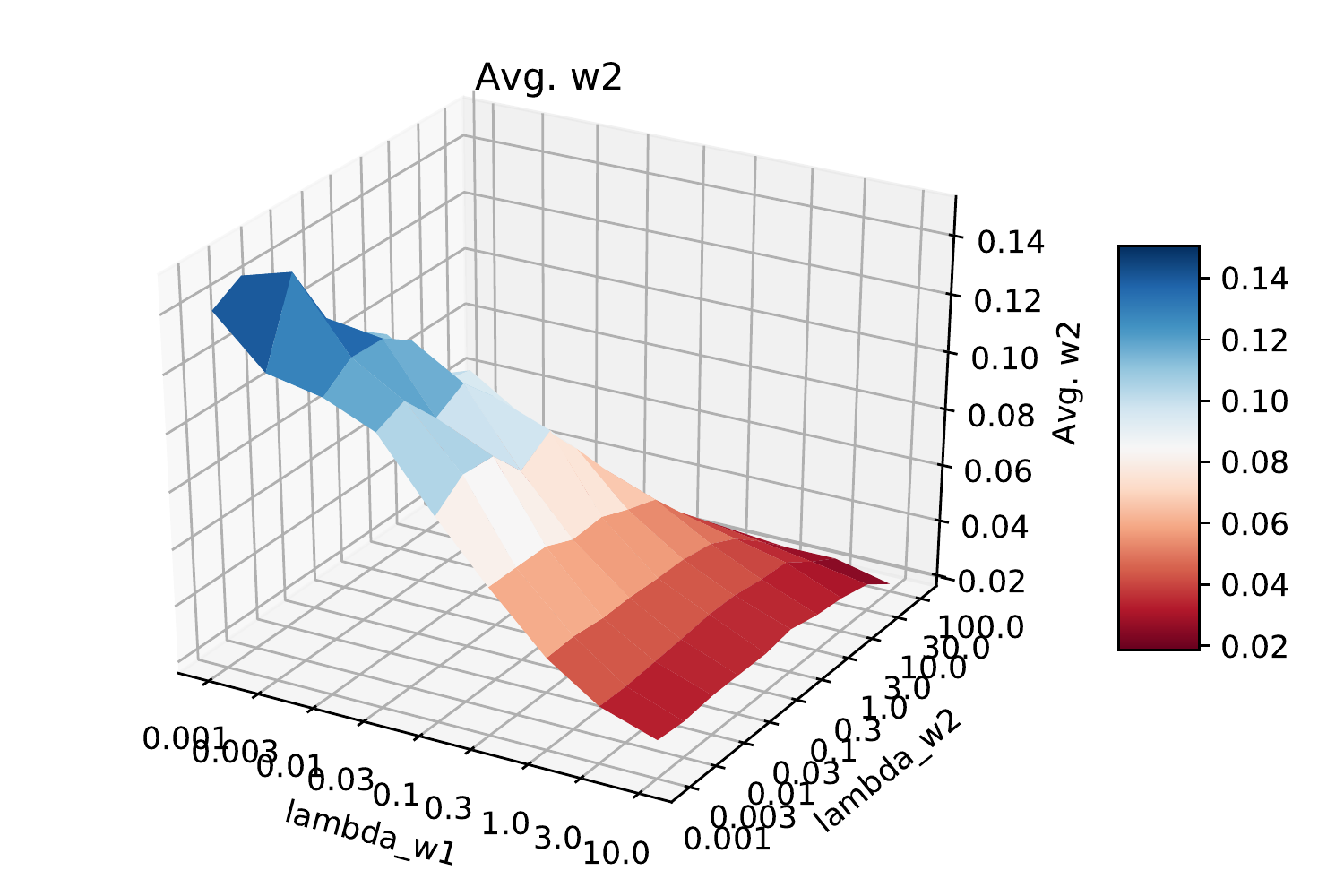}
	\end{subfigure}	
	~
	\begin{subfigure}[b]{0.28\textwidth}
		\includegraphics[width=\textwidth]{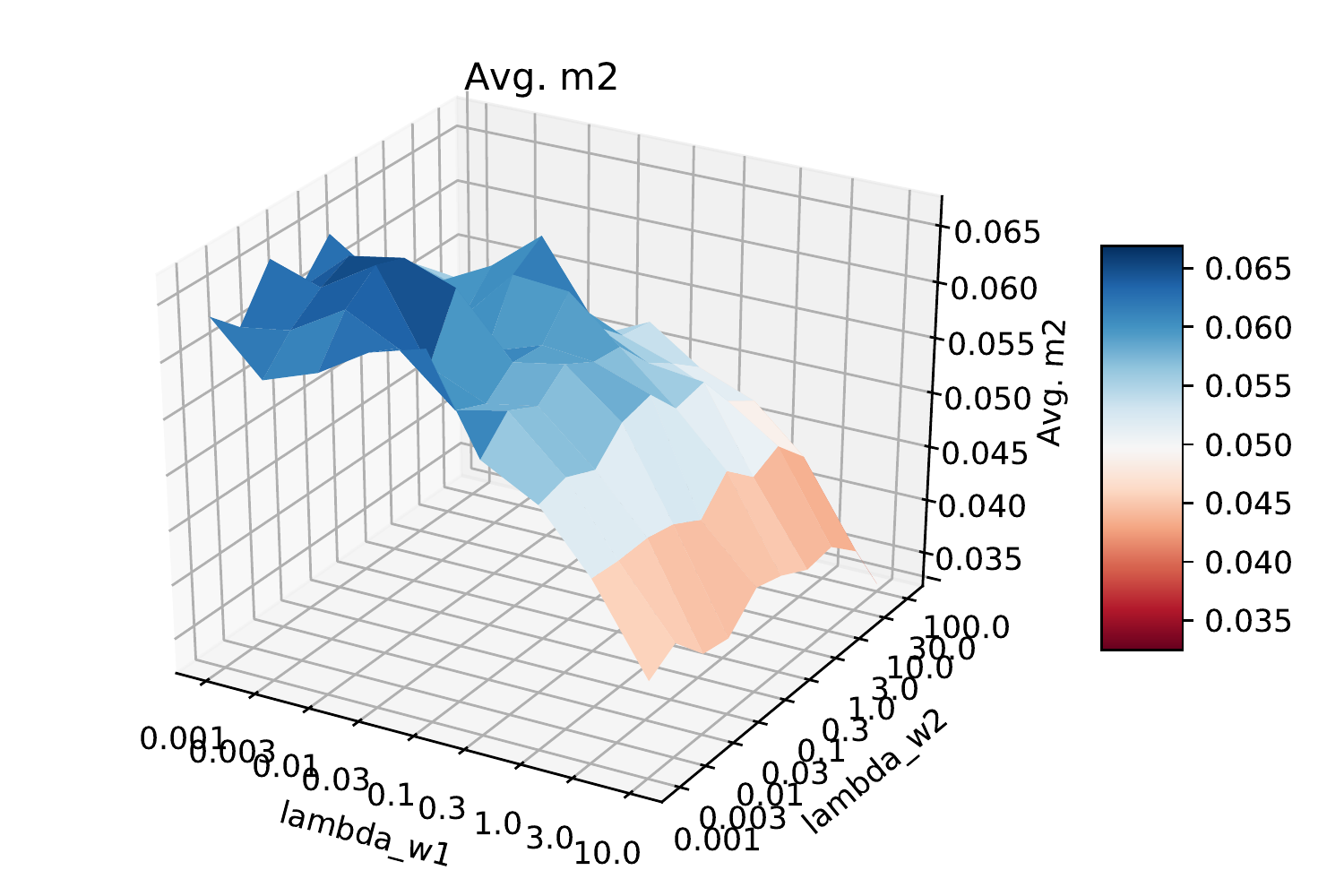}
	\end{subfigure}		
	\caption{3-D mesh surface plots of AUC(N), AUC(C), $w_1$, $w_2$, $m_1$ and $m_2$ of different combinations of $\lambda_{w_1}$ and $\lambda_{w_2}$, with a fixed $\lambda_r = 0.1$ from the ASSIST2009 dataset.}
	\label{fig:lw1lw2_mesh_result}
\end{figure*}

The experiment results are shown in Table~\ref{table:dkt_best_result} which gives a comparison of DKT models with and without regularization with respect to all of the evaluation measures. For clarity, here the DKT model without regularization is simply denoted as DKT, while the DKT model with regularization is denoted as DKT+. 

For the ASSIST2009 dataset, the DKT achieves an average test AUC(N) of 0.8212, while the DKT+ performs slightly better with an AUC(N) of 0.8227. However, for the DKT+, there is a considerable improvement in AUC(C) with an increase from 0.9044 to 0.9625. The waviness quantities also decrease significantly, from 0.0830 to 0.0229 for $w_1$ and from 0.1279 to 0.0491 for $w_2$. Moreover, although the DKT has already made half of the predictions change in the right direction, the DKT+ further uplifts the values of $m_1$ and $m_2$ from 0.3002 to 0.4486 and from 0.0156 to 0.0573, respectively.

Similar improvements in the evaluation measures are observed in ASSIST2015 as well. The DKT+ retains a similar AUC(N) of 0.7371, compared to that of the DKT. The values of AUC(C), $m_1$ and $m_2$ are boosted to 0.9233, 0.8122 and 0.0591, respectively. Moreover, the values of $w_1$ and $w_2$ in the DKT+ are only half of those for the DKT. 

As for ASSISTChall, although the performance of AUC(N) slightly decreases from 0.7343 to 0.7285 in the DKT+, improvement with respect to the other evaluation criteria is very significant. The DKT+ pushes the AUC(C) from 0.7109 to 0.8570 and reduces the $w_1$ from 0.0690 to 0.0147 and the $w_2$ from 0.1045 to 0.0301. Moreover, the DKT+ also improves the performance in $m_1$ and $m_2$, from 0.1151 to 0.3052 and from $-0.0055$ to 0.0441, respectively.

For statics2011, a noticeable increase is observed in both AUC(N) and AUC(C), from 0.8159 to 0.8349 and from 0.7404 to 0.9038, respectively. Moreover, $w_1$ and $w_2$ shrink from 0.1358 to 0.0074 and from 0.1849 to 0.0130, respectively. This substantial decrease in $w_1$ and $w_2$ would be ascribed to the large number of exercises in the dataset since the waviness regularizers act to confine the prediction changes on those exercises which are unrelated to the input. With a potentially substantial amount of unrelated exercises, $w_1$ and $w_2$ shrink significantly as a result. The DKT+ also ameliorates the situation that DKT makes more than half of the predictions change in the wrong direction. The values of $m_1$ and $m_2$ surge from $-0.25952$ to 0.47597 and from $-0.05090$ to 0.05712, respectively. 

With regard to Simulated-5, the DKT and the DKT+ result in a similar AUC(N), of 0.8252 and 0.8264, respectively. However, the DKT+ gives a huge improvement in AUC(C), $m_1$ and $m_2$. The DKT+ boosts the values of AUC(C) from 0.8642 to 0.9987, $m_1$ from $-0.1512$ to 0.9064 and $m_2$ from $-0.0134$ to $0.1659$. This means the DKT+ model makes almost all of the predictions and the prediction changes for the input exercise correct. Moreover, the waviness in the prediction transition is also reduced.

In summary, the experiment results reveal that the regularization based on $r$, $w_1$ and $w_2$ effectively alleviates the reconstruction problem and the wavy transition in prediction without sacrificing the prediction accuracy for the next interaction. Furthermore, for some combinations of $\lambda_r$, $\lambda_{w_1}$ and $\lambda_{w_2}$, the DKT+ even slightly outperforms the DKT in AUC(N).

\section{Discussion}
Apart from the experiment results, we plot Figures~\ref{fig:dkt_lambda_r_on_aucc} and~\ref{fig:lw1lw2_mesh_result} to better understand how the regularizers based on reconstruction and waviness affect the performance with respect to different evaluation measures. 

In Figure~\ref{fig:dkt_lambda_r_on_aucc}, we plot the average test AUC(N) and AUC(C) of different values of $\lr$ over all combinations of $\lambda_{w_1}$ and $\lambda_{w_2}$. It is observed that the higher the $\lr$ is, the higher the AUC(C) is achieved for all of the datasets. On the other hand, the AUC(N) generally decreases when the $\lr$ increases, but the downgrade in AUC(N) is not significant compared with the upgrade in AUC(C). This reveals that the reconstruction regularizer $r$ robustly resolves the reconstruction problem, without sacrificing much of the performance in AUC(N). Moreover, from the result in Table \ref{table:dkt_best_result}, we are usually able to find a combination of $\lr$, $\lambda_{w_1}$ and $\lambda_{w_2}$ that gives a comparable or even better AUC(N). This implies that the waviness regularizers can help to mitigate the slight degradation in AUC(N) incurred by the reconstruction regularizer.

To also see how the regularization parameters $\lambda_{w_1}$ and $\lambda_{w_2}$ affect the evaluation measures, their 3-D mesh plots, with a fixed $\lr=0.1$, for the ASSIST2009 dataset are shown in Figure~\ref{fig:lw1lw2_mesh_result}. The AUC(N) has a relatively flat and smooth surface when $\lambda_{w_1}$ lies between 0.0 and 1.0 and $\lambda_{w_2}$ lies between 0.0 and 10.0. Within this region, the DKT+ model also results in a higher AUC(C) value between 0.94 and 0.96. The performance of AUC(N) and AUC(C) starts to decline when $\lambda_{w_1}$ and $\lambda_{w_2}$ are larger than 1.0 and 10.0, respectively. It suggests that the model performance has a low sensitivity in AUC(N) and AUC(C) with respect to the hyperparameters $\lambda_{w_1}$ and $\lambda_{w_2}$. As for the waviness measures $w_1$ and $w_2$, they decrease in a bell-like shape when $\lambda_{w_1}$ and $\lambda_{w_2}$ increase. With regard to the consistency measures, even though the mesh surface is a bit bumpy, $m_1$ increases along with larger values of $\lambda_{w_1}$ and $\lambda_{w_2}$ within the same range mentioned above. This observation implies that both the reconstruction regularizer and the waviness regularizers help to improve the prediction consistency for the current input. On the other hand, $m_2$ has a decreasing trend with larger values of $\lambda_{w_1}$ and $\lambda_{w_2}$. This is reasonable because the waviness regularizers will reduce the prediction change between the outputs and thus the value of $m_2$, which takes the change in magnitude into account, is reduced. All in all, the robustness of the regularizers $w_1$ and $w_2$ is ascertained thanks to the low sensitivity in the prediction accuracy (AUC(N) and AUC(C)), the observable reduction in the waviness measures ($w_1$ and $w_2$), and the improvement in the consistency measures ($m_1$ and $m_2$).

\begin{figure*}[!b]
	\centering
	\begin{subfigure}[b]{0.9\textwidth}
		\includegraphics[width=\textwidth]{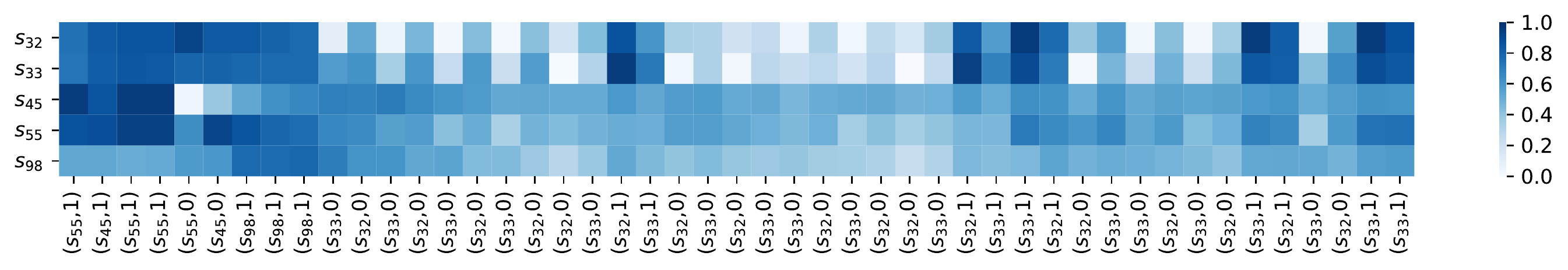}
		\includegraphics[width=\textwidth]{dkt_id1.pdf}
		\caption{Heatmap (upper: DKT+; lower: DKT) for the predicted probability of correctly answering each skill answered by the student. }
		\label{fig:dkt_id1}	
        
        \vfill
        
		\includegraphics[width=\textwidth]{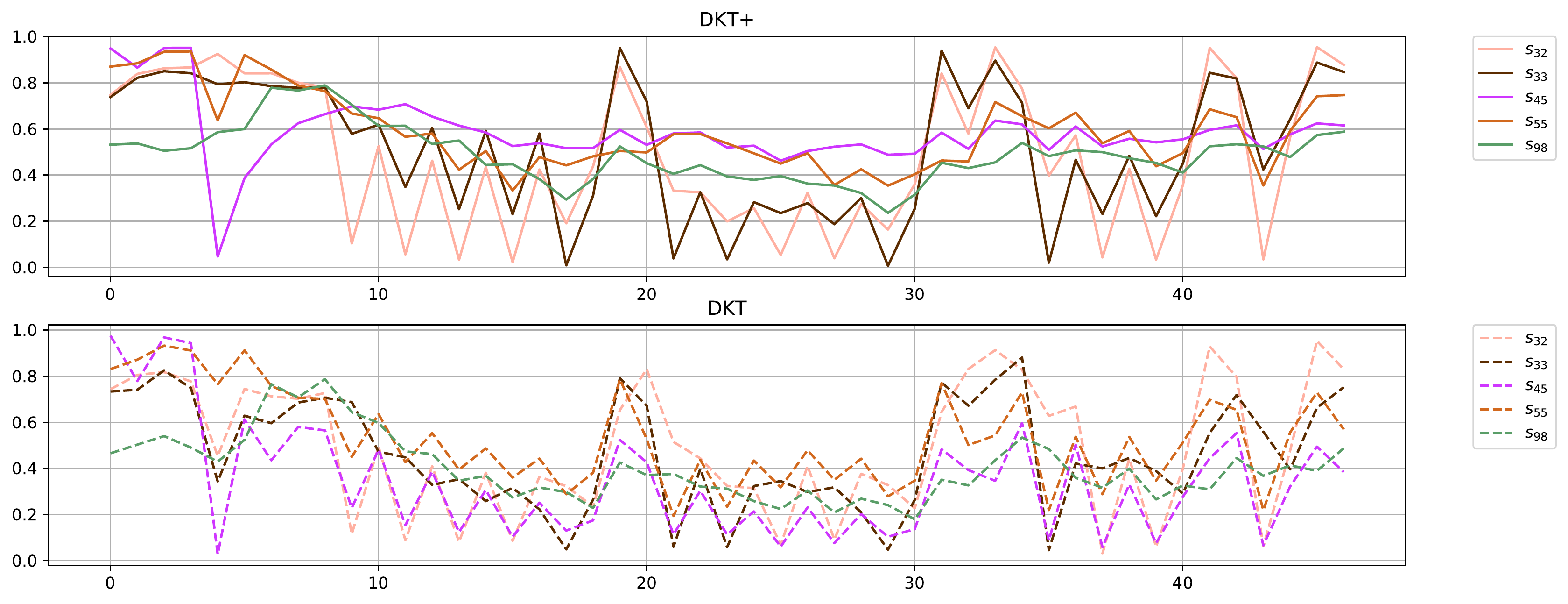}
		\caption{Line plot (upper: DKT+; lower: DKT) for the predicted probability of correctly answering each skill answered by the student. It aims to show the directional change in prediction of each skill in the DKT/DKT+ model.}
		\label{fig:comparison_id1}	
        
        \vfill
        
		\includegraphics[width=\textwidth]{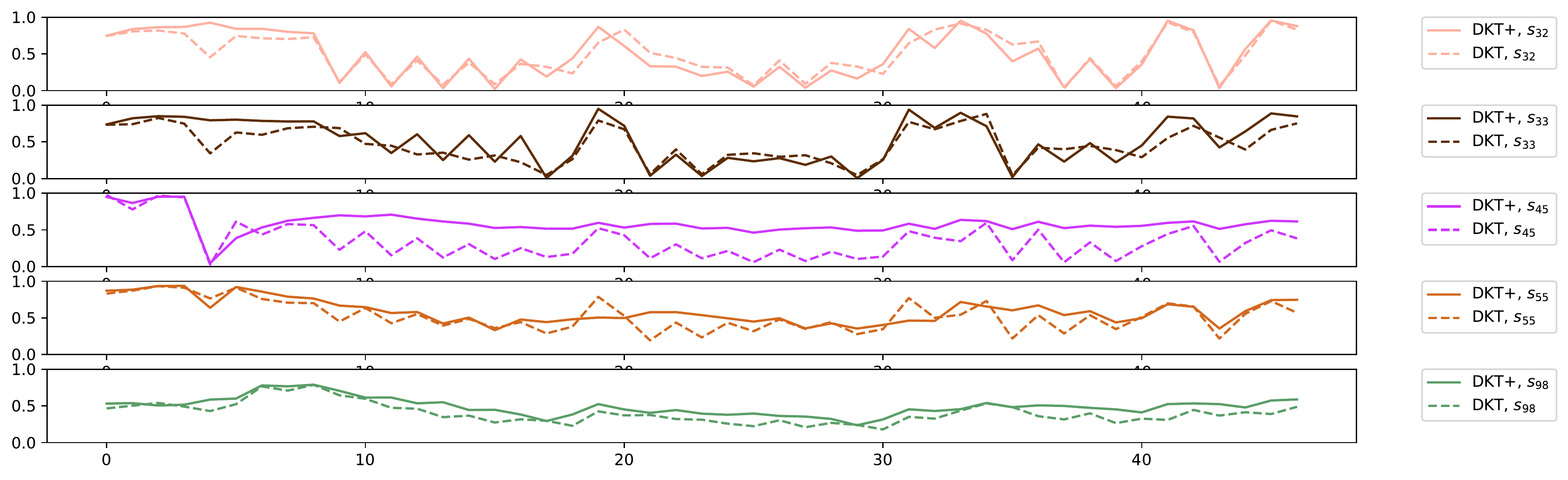}
		\caption{Line plot for the each skill's prediction, which is visualized separately with regard to the skill tag. It aims to compare the prediction result of the same skill between DKT and DKT+.}
		\label{fig:id1_exercise_plot}
	\end{subfigure}
    
	\caption{Visualization of a subset of DKT's output layer using the student's interaction sequences (id-1) extracted from ASSIST2009. The DKT+ model used is trained with $\lr=0.10$, $\lambda_{w_1}=0.003$, $\lambda_{w_2}=3.0$. We note that $s_{32}$ is ``Ordering Positive Decimals'', $s_{33}$ is ``Ordering Fractions'', $s_{45}$ is ``Subtraction Whole Numbers'', $s_{55}$ is ``Absolute Value'', and $s_{98}$ is ``Equation Solving Two or Fewer Steps''.}
	\label{fig:dkt_visualization_id1}
\end{figure*}

In addition to the overall goodness in terms of the evaluation measures, the prediction results of DKT and DKT+ for the student (id-1) are visualized in Figure~\ref{fig:dkt_visualization_id1} in order to give a better sense of the regularization's effect on the prediction. Specifically, the prediction results are visualized in two line plots, in addition to the heatmap plot. The first line plot illustrates the change in prediction of all the answered skills of the DKT/DKT+ model, while the second line plot emphasizes the change in prediction of each skill between DKT and DKT+, showing their differences in prediction. Concretely, as for the DKT, it shows a relatively wavy transition of knowledge state across time-steps (Figure~\ref{fig:dkt_id1}). Moreover, the predicted knowledge states of most of the skills share the same directional change in prediction in DKT (Figure~\ref{fig:comparison_id1}). This means that when a student answers a question wrongly, most of the predicted skills' mastery level will decrease simultaneously, and vice versa. However, it is intuitively untrue, as answering skill $s_i$ wrongly does not necessarily lead to a knowledge fade in other skills. On the other hand, the DKT+ demonstrates a smooth prediction transition notably. For example, as seen from Figures~\ref{fig:dkt_id1} and \ref{fig:id1_exercise_plot}, when the DKT+ receives the inputs $(s_{32}, 0)$ or $(s_{33}, 0)$, the changes in prediction for $s_{45}$, $s_{55}$ and $s_{98}$ across time-steps are not as wavy as those in the DKT, revealing that DKT+ retains latent knowledge state in RNN for $s_{45}$, $s_{55}$ and $s_{98}$ from the previous time-steps. This consistent prediction will alleviate the misinterpretation of the student knowledge state caused by the wavy transition problem, and enhance the interpretability of the knowledge state in DKT.


\section{Conclusion and future work}
This paper points out two problems which arise when interpreting DKT's prediction results: (1)~the reconstruction problem, and (2)~the wavy transition in prediction. Both problems are undesirable because it would mislead the interpretation of students' knowledge states. We thereby proposed three regularization terms for enhancing the consistency of prediction in DKT. One of them is the reconstruction error $r$, which is evaluated in AUC(C), $m_1$ and $m_2$. The other two are waviness measures $w_1$ and $w_2$, which are the norms for measuring the changes between two consecutive prediction output vectors and are used directly as evaluation measures. Experiments show that these regularizers effectively alleviate the two problems without sacrificing the prediction accuracy (AUC(N)) on the original task which is to predict the next interaction performance.

Although the reconstruction regularizer improves the performance with respect to AUC(C) and the waviness regularizers reduce the waviness in the prediction transition, it is hard to say how low the values of $w_1$ and $w_2$ should be in order to qualify for a good KT model. Ideally, a KT model should only change those prediction components which are related to the current input while keeping the other components unchanged or only changed slightly due to some other subtle reasons, e.g., forgetting. Nevertheless, the KC-dependency graphs are different from one dataset to another dataset, so different values of $w_1$ and $w_2$ are expected in their ideal KT models.

Moreover, more work should be done on improving the stability and accuracy of the prediction for unseen data, more specifically the unobserved KCs. The objective function and the evaluation measures for the DKT+ take only the current and next interactions into account. There is no consideration for interactions in the further future, let alone the evaluation measures for the prediction precision of the unobserved KCs. Yet, unobserved KCs are of vital importance because an ITS should make personalized recommendation on the learning materials for students over not only the observed KCs, but also the unobserved ones. An accurate estimation on the unobserved KCs will help an ITS provide more intelligent pedagogical guidance to students. One of the possible extensions of this work is to take the further future interaction into account when training the DKT model:
\begin{equation} \label{new_objective_function2}
\mathcal{L} =  
\frac
{1}
{c}
\left( 
  \sum_{i=1}^{n} 
  \sum_{t=1}^{T_i - 1}
  \sum_{j=1}^{T_i - t}
  \gamma^{j-1} l \left( \y_{t}^{i} \cdot \deltabm (q_{t+j}), a_{t+j} \right) 
\right)
\end{equation}
where $c = \sum_{i=1}^{n} \sum_{t=1}^{T_i - 1} \sum_{j=1}^{T_i - t} \gamma^{j-1}$ is the normalizing term, $\gamma$ is the decay factor similar to that in reinforcement learning. This potentially leads the DKT model to learn a more robust representation of the latent knowledge state.

\section{Acknowledgments}
This research has been supported by the project ITS/205/15FP from the Innovation and Technology Fund  of Hong Kong.

\clearpage
\balance{}

\bibliographystyle{SIGCHI-Reference-Format}
\bibliography{reference}


\begin{thebibliography}{00}


\ifx \showCODEN    \undefined \def \showCODEN     #1{\unskip}     \fi
\ifx \showDOI      \undefined \def \showDOI       #1{{\tt DOI:}\penalty0{#1}\ }
  \fi
\ifx \showISBNx    \undefined \def \showISBNx     #1{\unskip}     \fi
\ifx \showISBNxiii \undefined \def \showISBNxiii  #1{\unskip}     \fi
\ifx \showISSN     \undefined \def \showISSN      #1{\unskip}     \fi
\ifx \showLCCN     \undefined \def \showLCCN      #1{\unskip}     \fi
\ifx \shownote     \undefined \def \shownote      #1{#1}          \fi
\ifx \showarticletitle \undefined \def \showarticletitle #1{#1}   \fi
\ifx \showURL      \undefined \def \showURL       #1{#1}          \fi

\bibitem{ITS2006_Cen_LFA}
{Hao Cen}, {Kenneth Koedinger}, {and} {Brian Junker}. 2006.
\newblock \showarticletitle{Learning factors analysis -- a general method for
  cognitive model evaluation and improvement}. In {\em Proceedings of the 8th
  International Conference in Intelligent Tutoring Systems}. Springer, Berlin,
  Heidelberg, 164--175.
\newblock


\bibitem{UMUAI1994_Corbett_BKT}
{Albert~T. Corbett} {and} {John~R. Anderson}. 1995.
\newblock \showarticletitle{Knowledge tracing: modeling the acquisition of
  procedural knowledge}.
\newblock {\em User Modeling and User-Adapted Interaction\/} {4}, 4 (March
  1995), 253--278.
\newblock


\bibitem{EDM2014_Gonzales_General}
{Jos{\'e} Gonz{\'a}lez-Brenes}, {Yun Huang}, {and} {Peter Brusilovsky}. 2014.
\newblock \showarticletitle{General features in knowledge tracing to model
  multiple subskills, temporal item response theory, and expert knowledge}. In
  {\em Proceedings of the 7th International Conference on Educational Data
  Mining}. 84--91.
\newblock


\bibitem{ITS2014_hawkins_learning}
{William~J. Hawkins}, {Neil~T. Heffernan}, {and} {Ryan~S.J.D. Baker}. 2014.
\newblock \showarticletitle{Learning Bayesian knowledge tracing parameters with
  a knowledge heuristic and empirical probabilities}. In {\em Proceedings of
  the 12th International Conference on Intelligent Tutoring Systems}. Springer,
  Cham, 150--155.
\newblock


\bibitem{IEEE2017_Kaesor_DynamicBKT}
{Tanja Kaeser}, {Severin Klingler}, {Alexander~G. Schwing}, {and} {Markus
  Gross}. 2017.
\newblock \showarticletitle{Dynamic Bayesian networks for student modeling}.
\newblock {\em IEEE Transactions on Learning Technologies\/}  {10} (March
  2017), 450--462.
\newblock


\bibitem{EDM2016_Khajah_How}
{Mohammad Khajah}, {Robert~V. Lindsey}, {and} {Michael~C. Mozer}. 2016.
\newblock \showarticletitle{How deep is knowledge tracing}. In {\em Proceedings
  of the 9th International Conference on Educational Data Mining}. 94--101.
\newblock


\bibitem{ARXIV2015_Lipton_RNN}
{Zachary~C. Lipton}, {John Berkowitz}, {and} {Charles Elkan}. 2015.
\newblock \showarticletitle{A critical review of recurrent neural networks for
  sequence learning}.
\newblock {\em ArXiv e-prints 1506.00019\/} (May 2015).
\newblock


\bibitem{olah2015understanding}
{Christopher Olah}. 2015.
\newblock Understanding LSTM networks.
\newblock Colah. github. io.   (Augest 2015).
\newblock
\newblock
\shownote{Retrieved December 10, 2017 from
  \url{http://colah.github.io/posts/2015-08-Understanding-LSTMs/}.}


\bibitem{UMAP2010_Pardos_BKT}
{Zachary~A. Pardos} {and} {Neil~T. Heffernan}. 2010.
\newblock \showarticletitle{Modeling individualization in a Bayesian networks
  implementation of knowledge tracing}. In {\em Proceedings of the 18th
  International Conference on User Modeling, Adaptation, and Personalization},
  Vol. 6075. Springer, Berlin, Heidelberg, 255--266.
\newblock


\bibitem{UMAP2011_Pardos_KT}
{Zachary~A. Pardos} {and} {Neil~T. Heffernan}. 2011.
\newblock \showarticletitle{KT-IDEM: introducing item difficulty to the
  knowledge tracing model}. In {\em Proceedings of the 19th International
  Conference on User Modeling, Adaption and Personalization}, Vol. 6787.
  Springer, 243--254.
\newblock


\bibitem{AIE2009_Pavlik_PFA}
{Philip~I. Pavlik}, {Hao Cen}, {and} {Kenneth~R. Koedinger}. 2009.
\newblock \showarticletitle{Performance factors analysis -- a new alternative
  to knowledge tracing}. In {\em Proceedings of the 14th International
  Conference on Artificial Intelligence in Education}. IOS Press, Amsterdam,
  Netherlands, 531--538.
\newblock


\bibitem{NIPS2015_Piech_DKT}
{Chris Piech}, {Jonathan Bassen}, {Jonathan Huang}, {Surya Ganguli}, {Mehran
  Sahami}, {Leonidas~J Guibas}, {and} {Jascha Sohl-Dickstein}. 2015.
\newblock \showarticletitle{Deep knowledge tracing}. In {\em Advances in Neural
  Information Processing Systems}. 505--513.
\newblock


\bibitem{NN1998_Prehelt_EarlyStopping}
{Lutz Prechelt}. 1998.
\newblock \showarticletitle{Automatic early stopping using cross validation:
  quantifying the criteria}.
\newblock {\em Neural Networks\/} {11}, 4 (June 1998), 761--767.
\newblock


\bibitem{L@S2016_Reddy_LSE}
{Siddharth Reddy}, {Igor Labutov}, {and} {Thorsten Joachims}. 2016.
\newblock \showarticletitle{Learning student and content embeddings for
  personalized lesson Sequence Recommendation}. In {\em Proceedings of the 3rd
  ACM Conference on Learning @ Scale}. ACM, New York, NY, USA, 93--96.
\newblock


\bibitem{arxiv2016_Steven_Modeling}
{Steven Tang}, {Joshua~C. Peterson}, {and} {Zachary~A. {Pardos}}. 2016.
\newblock \showarticletitle{Modelling student behavior using granular large
  scale action data from a MOOC}.
\newblock {\em ArXiv e-prints 1608.04789\/} (Augest 2016).
\newblock


\bibitem{EDM2017_Wang_Programming}
{Lisa Wang}, {Angela Sy}, {Larry Liu}, {and} {Chris Piech}. 2017.
\newblock \showarticletitle{Learning to represent student knowledge on
  programming exercises using deep learning}. In {\em Proceedings of the 10th
  International Conference on Educational Data Mining}. 324--329.
\newblock


\bibitem{IAIED2013_Wang_Extending}
{Yutao Wang} {and} {Neil~T. Heffernan}. 2013.
\newblock \showarticletitle{Extending knowledge tracing to allow partial
  credit: using continuous versus binary nodes}. In {\em Proceedings of the
  13th International Conference on Artificial Intelligence in Education}.
  Springer, Berlin, Heidelberg, 181--188.
\newblock


\bibitem{EDM2016_Wilson_IRT}
{Kevin~H. Wilson}, {Yan Karklin}, {Bojian Han}, {and} {Chaitanya Ekanadham}.
  2016.
\newblock \showarticletitle{Back to the basics: Bayesian extensions of IRT
  outperform neural networks for proficiency estimation}. In {\em Proceedings
  of the 9th International Conference on Educational Data Mining}. 539--544.
\newblock


\bibitem{EDM2016_Xiong_Going}
{Xiaolu Xiong}, {Siyuan Zhao}, {Eric Van~Inwegen}, {and} {Joseph Beck}. 2016.
\newblock \showarticletitle{Going deeper with deep knowledge tracing}. In {\em
  Proceedings of 9th International Conference on Educational Data Mining}.
  545--550.
\newblock


\bibitem{IAIED2013_Yudelson_individualized}
{Michael~V. Yudelson}, {Kenneth~R. Koedinger}, {and} {Geoffrey~J. Gordon}.
  2013.
\newblock \showarticletitle{Individualized Bayesian knowledge tracing models}.
  In {\em Proceedings of the 16th International Conference on Artificial
  Intelligence in Education}, Vol. 7926. Springer, Berlin, Heidelberg,
  171--180.
\newblock


\bibitem{WWW2017_Zhang_DKVMN}
{Jiani Zhang}, {Xingjian Shi}, {Irwin King}, {and} {Dit-Yan Yeung}. 2017.
\newblock \showarticletitle{Dynamic key-value memory networks for knowledge
  tracing}. In {\em Proceedings of the 26th International Conference on World
  Wide Web}. International World Wide Web Conferences Steering Committee,
  Republic and Canton of Geneva, Switzerland, 765--774.
\newblock


\bibitem{JRSS2005_Zou_ElasticNet}
{Hui Zou} {and} {Trevor Hastie}. 2004.
\newblock \showarticletitle{Regularization and variable selection via the
  elastic net}.
\newblock {\em Journal of the Royal Statistical Society: Series B (Statistical
  Methodology)\/} {67}, 2 (September 2004), 301--320.
\newblock


\end{thebibliography}

\end{document}